\documentclass[accepted]{melba}

\usepackage{amsmath,amsfonts}

\usepackage{algorithm, algorithmic}
\usepackage{multirow, multicol, booktabs, microtype}
\usepackage{enumitem}

\usepackage{tabularx}
\usepackage{pgfplots}
\usepackage{tikz}
\usepackage{needspace}

\hypersetup{
  colorlinks,
  citecolor=[rgb]{0.0,0.7,0.5},
  filecolor=[rgb]{0.6,0.0,0.0},
  linkcolor=[rgb]{0.0,0.7,0.5},
  urlcolor=[rgb]{0.0,0.0,0.5}
}

\melbaid{2026:002}
\doi{10.59275/j.melba.2026-3cg7}
\melbaauthors{Tyler Ward, Aaron Moseley, and Abdullah-Al-Zubaer Imran}
\email{aimran@uky.edu}
\volume{2026}
\firstpageno{21}
\melbayear{2026}
\datesubmitted{2025-05-25}
\datepublished{2026-03-13}

\ShortHeadings{Domain and Task-Focused Example Selection for Contrastive Medical Segmentation}{Ward et al.}

\title{Domain and Task-Focused Example Selection for Data-Efficient Contrastive Medical Image Segmentation}

\author{
	\firstname Tyler \surname Ward \orcid{0000-0003-0669-1407},
	\firstname Aaron \surname Moseley, 
        \firstname Abdullah-Al-Zubaer \surname Imran \orcid{0000-0001-5215-339X}
}

\affiliations{
\addr Department of Computer Science, University of Kentucky, Lexington, KY 40506, USA
}

\abstract{
Segmentation is one of the most important tasks in the medical imaging pipeline as it influences a number of image-based decisions. To be effective, fully supervised segmentation approaches require large amounts of manually annotated training data. However, the pixel-level annotation process is expensive, time-consuming, and error-prone, hindering progress and making it challenging to perform effective segmentations. Therefore, models must learn efficiently from limited labeled data. Self-supervised learning (SSL), particularly contrastive learning via pre-training on unlabeled data and fine-tuning on limited annotations, can facilitate such limited labeled image segmentation. To this end, we propose a \textit{novel} self-supervised contrastive learning framework for medical image segmentation, leveraging inherent relationships of different images, dubbed \textit{PolyCL}. Without requiring any pixel-level annotations or unreasonable data augmentations, our PolyCL learns and transfers context-aware discriminant features useful for segmentation from an innovative surrogate, in a task-related manner. Additionally, we integrate the Segment Anything Model (SAM) into our framework in two novel ways: as a post-processing refinement module that improves the accuracy of predicted masks using bounding box prompts derived from coarse outputs, and as a propagation mechanism via SAM 2 that generates volumetric segmentations from a single annotated 2D slice. Experimental evaluations on three public computed tomography (CT) datasets demonstrate that PolyCL outperforms fully-supervised and self-supervised baselines in both low-data and cross-domain scenarios.\\
Our code is available at~\url{https://github.com/tbwa233/PolyCL}.
}

\keywords{Computed tomography, contrastive learning, medical image segmentation, self-supervised learning, SAM}

\begin{document}

\twocolumn[\maketitle]

\section{Introduction}
\enluminure{M}edical image segmentation is essential in healthcare for clear interpretations of anatomical structures or lesions in images like computed tomography (CT), aiding accurate diagnoses and monitoring of health conditions. However, obtaining a fully annotated dataset is expensive as it requires expert radiologists to label each pixel of a structure. This costliness limits the effectiveness of segmentation models, impeding their adoption in healthcare. Therefore, ongoing research is striving to optimize data usage, enabling models to effectively perform image segmentation even with reduced data \cite{peng2021medical, chaitanya2023local, imran2020fully}. Deviating from its supervised learning counterparts, self-supervised learning (SSL) has become an attractive alternative to this end, e.g., \cite{shurrab2022self}. In SSL, a model is first pre-trained on unlabeled samples to learn a pretext task and then fine-tuned on labeled samples to learn a downstream task for the actual evaluation. The representation learned by the model from the pre-training stage through a surrogate supervision (labels created from data itself) can be successfully transferred to various downstream tasks, including segmentation \cite{chaitanya2023local, imran2020self}.  

\begin{figure}[t]
    \centering
    \includegraphics[width=\linewidth, trim={0cm 0cm 0cm 0cm},clip]{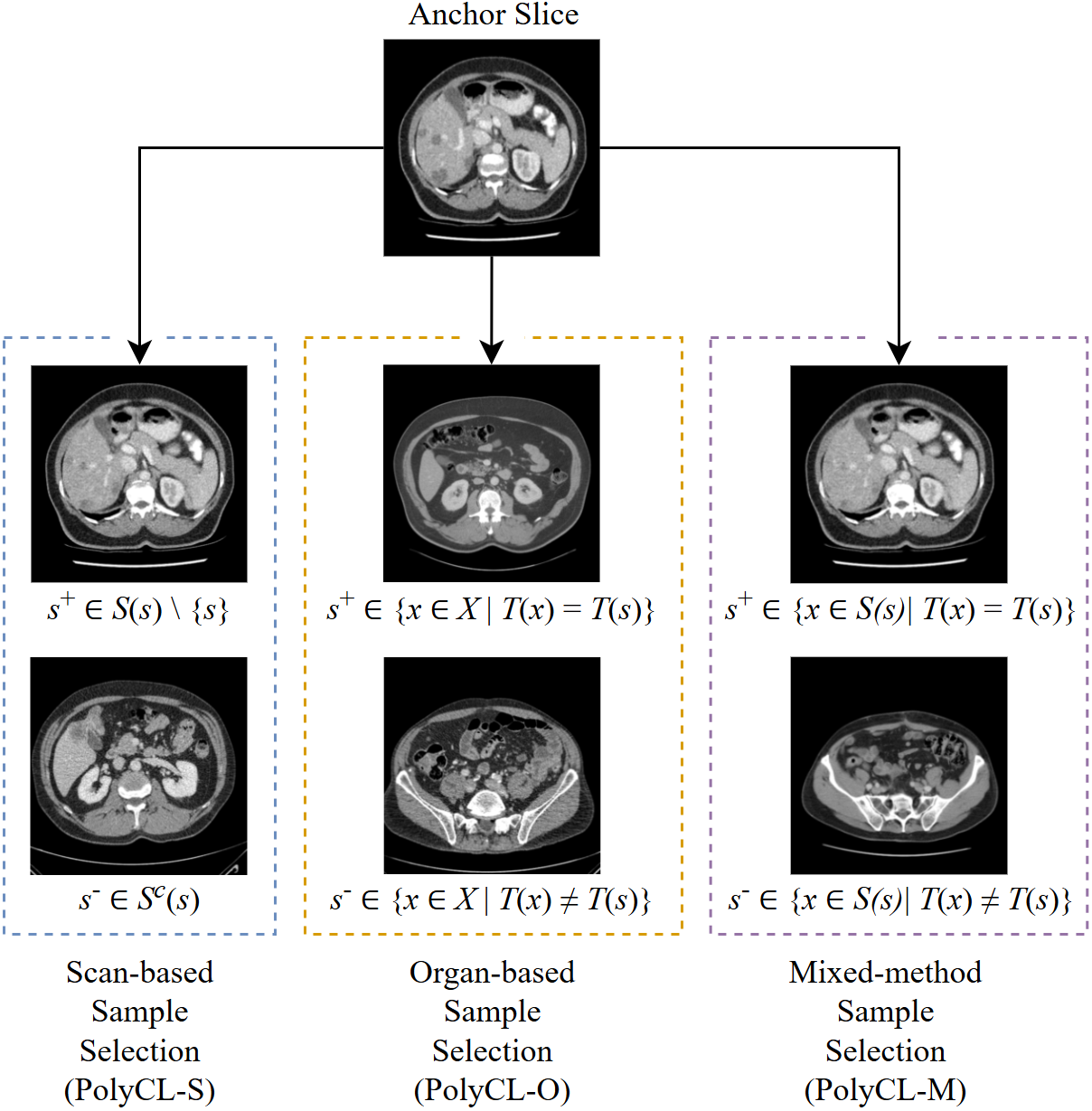}
    \caption{Example selection strategies for the proposed PolyCL framework: PolyCL-S uses the information of the CT scan to which each slice belongs, while PolyCL-O uses the information of whether each slice contains the target organ-of-interest; PolyCL-M uses both the organ-based and scan-based information.}
    \label{fig:updated_teaser}
\end{figure}

A successful variant of self-supervised learning is contrastive learning, which, as a pre-training strategy, helps cluster unlabeled examples in the latent space \cite{chen2020simple}. In contrastive learning-based pre-training, positive and negative image pairs are created for each slice in a dataset. A model is trained on aligning representations by increasing the similarity between positive pairs and increasing the difference between negative pairs in the embedding space. Most contrastive pre-training strategies use data augmentations to create positive examples, encouraging the model to represent transformed versions of the same image similarly \cite{chen2020simple, yan2022sam}. 

While it has shown benefits, this method fails to learn positive relationships between different images in the dataset. Previous work has investigated learning inter-scan relationships when looking at CT data, choosing positive example slices from similar locations within different scans \cite{xiang2021self}. This is less data-hungry than standard supervised segmentation training, but requires that the entire training dataset be obtained using the same modality, as the images must be correctly aligned. Other works have also looked at training the encoder to understand different views of the same subject or image \cite{tian2020contrastive, azizi2021big}. But these methods have not yet been adapted to medical images such as CT scans.

As mentioned, most previous work relies heavily on large pools of transformations to create positive examples for contrastive learning. Because of this, a major gap in the medical imaging field is contrastive pre-training strategies that leverage useful downstream task-related surrogates, enabling the model to learn useful information about the target task before seeing labeled data for fine-tuning. Our proposed framework, PolyCL, introduces such a strategy through organ-based and scan-based example selection strategies. Our specific contributions are summarized as follows:

\begin{itemize}
    \item A novel self-supervised contrastive learning-based pre-training approach for medical image segmentation.
    \item Innovative example-selection strategies leveraging different amounts of labeled data.
    \item Thorough experimentation demonstrating PolyCL's effectiveness and generalizability across multiple CT datasets.
    \item Incorporation of the Segment Anything model (SAM) as a post-finetuning method for mask refinement and propagation.
\end{itemize}

\begin{figure}[t]
    \centering
    \includegraphics[width=0.75\linewidth, trim={0cm 0cm 0cm 0cm},clip]{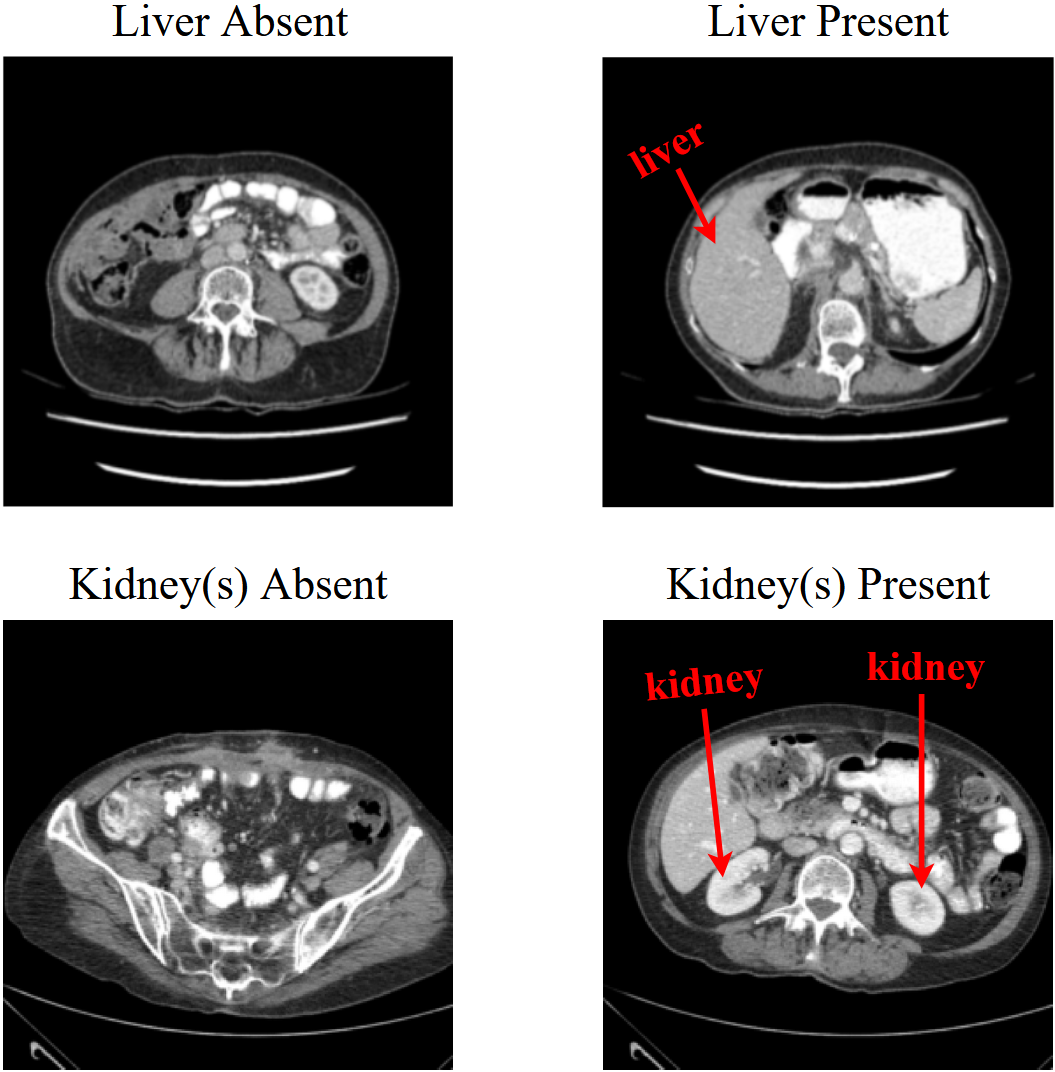}
    \caption{Sample abdominal CT images slices illustrating the presence and absence of liver/kidney(s).}
    \label{fig:liverappearance}
\end{figure}

\paragraph{Previous Work:}
This manuscript is an extension of our paper ``PolyCL: Context-Aware Contrastive Learning for Image Segmentation'' presented at the 2024 IEEE International Symposium on Biomedical Imaging (ISBI 2024)~\cite{moseley2024polycl}. This manuscript substantially extends by adding a thorough literature review, experiments with larger datasets, additional experiments and baselines, improved results, a more detailed description of the methods and results discussion, and additional figures and visualizations. The largest addition comes from our modified architecture. 

Our ISBI paper primarily focused on organ-based (PolyCL-O) and scan-based (PolyCL-S) example selection strategies. Here, we introduce a new example selection strategy combining these previous approaches for improved results (PolyCL-M). Additionally, we incorporate SAM in two innovative ways. The first method is applying it as a post-processing method, where it serves to refine roughly predicted segmentation masks produced from limited training data. The second method utilizes the recursive mask propagation capabilities of SAM 2 to extend 2D segmentation masks across an entire 3D CT volume, requiring as little as a single annotated slice.

\needspace{4\baselineskip}
\section{Related Work}

\subsection{Contrastive Learning}
In contrastive learning, common features are learned between similar instances, and different features are distinguished among dissimilar instances~\cite{chen2020simple}. Despite developing more slowly in comparison to its applications in natural language processing, contrastive learning approaches for computer vision have gained a lot of attention in recent years \cite{hojjati2024self}. Unlike fully-supervised methods, contrastive learning relies on identifying relative similarities between data points rather than absolute labels. This effectively reduces the substantial manual effort and computational resources required for fully-supervised training~\cite{hu2024comprehensive}. Momentum contrast (MoCo), first proposed by \cite{he2020momentum}, was one of the seminal works in making contrastive learning more memory efficient for computer vision. Unlike prior approaches that relied on large batch sizes to generate a sufficient amount of negative examples, MoCo maintains a queue of encoded samples from prior batches, updated using a momentum encoder. This decoupling of the encoder updates for current and past representations allows MoCo to scale efficiently while maintaining temporal consistency. 

Additional contrastive learning methods in computer vision shortly followed MoCo, such as SimCLR~\cite{chen2020simple}, simplified and popularized contrastive learning by focusing on data augmentations to create positive pairs. Through transformations such as cropping, rotation, and color distortion, SimCLR encourages models to align representations of augmented views of the same image while distinguishing them from representations of other images. It also introduced a nonlinear projection head to optimize the latent space, further enhancing the quality of learned features. While highly effective, SimCLR’s reliance on large batch sizes and extensive augmentations posed challenges for applications in resource-constrained environments. 

The supervised contrastive learning (SupCon) framework~\cite{khosla2020supervised} was built upon the foundation of SimCLR and addressed this issue by defining positive pairs based on shared class labels rather than augmentations, enabling alignment of semantically similar samples in the latent space. Another model, Bootstrap Your Own Latent (BYOL)~\cite{grill2020bootstrap}, represents a departure from the reliance on negative examples through the use of an online network to learn representations and a target network to provide stable targets for the learning process. The target network’s parameters are updated using an exponential moving average of the online network’s weights, ensuring stability without the need for explicit negative pairs.

Beyond these works, contrastive learning has witnessed significant innovation. Hybrid approaches now integrate contrastive objectives with other learning paradigms. For example,~\citet{xu2024deep} proposed a deep image clustering model combining contrastive learning with multi-scale graph convolutional neural networks. In another work, \citet{zhang2024skip} combined contrastive learning with predictive coding for time-series anomaly detection. Researchers have also explored contrastive learning for multi-modal data, enabling cross-domain representation alignment, such as combining visual features with information from other modalities~\cite{radford2021learning, imran2022multimodal}. These advancements reflect a growing emphasis on flexibility, robustness, and task-specific adaptation in contrastive learning frameworks.

One such task where contrastive learning has found many uses is in medical imaging, where its ability to leverage large unlabeled datasets has proven extremely useful. Unlike natural image datasets, where augmentations suffice to create meaningful positive pairs, medical datasets require strategies tailored to anatomical structures and modality-specific features~\cite{ma2024segment}. Techniques like inter-scan pairing, where positive examples are derived from similar anatomical locations across scans, have been employed to capture inter-patient consistency~\cite{salari2024camld}. Pseudo-labeling\cite{chen2024apan} and anatomical embeddings~\cite{yu2024drasclr} have also been explored to align representations of clinically similar cases while maintaining task relevance.

\begin{figure*}
    \centering
    \includegraphics[width=\linewidth, trim={0cm 0.0cm 0cm 0cm},clip]{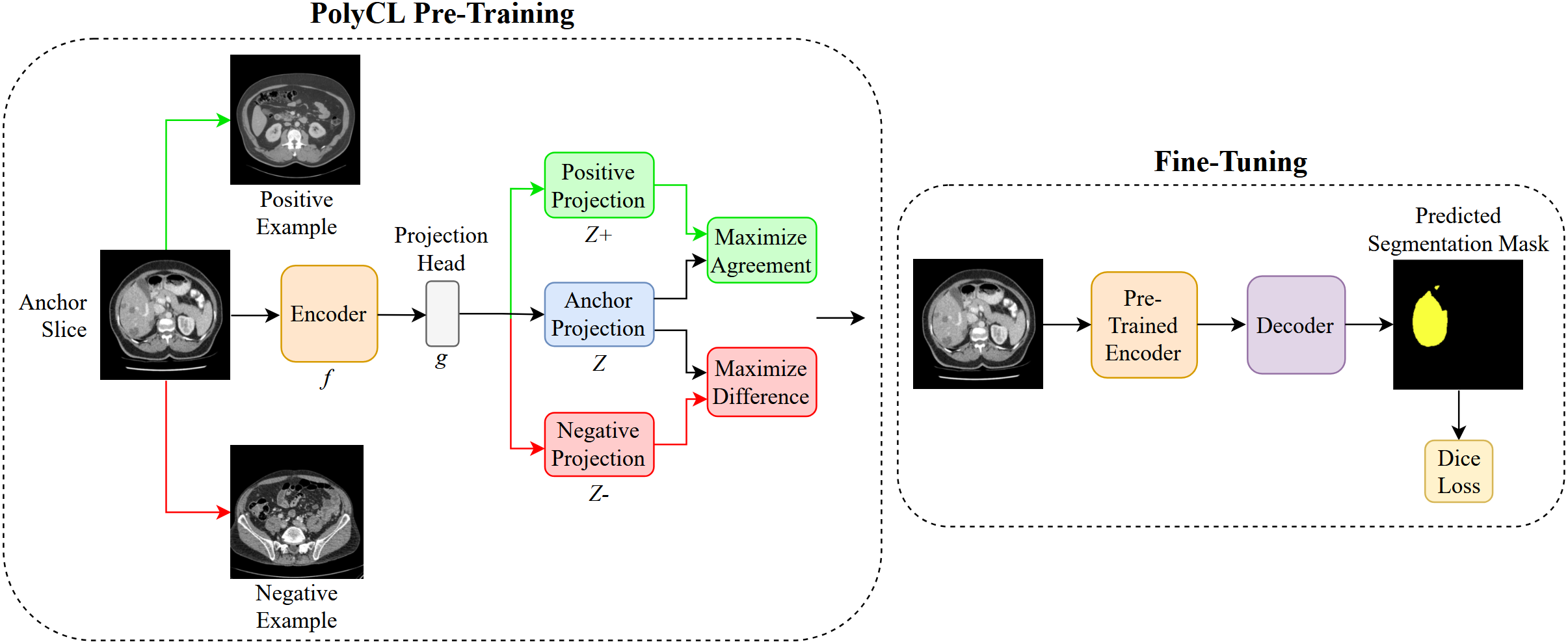}
    \caption{Overview of the proposed PolyCL training framework for medical image segmentation. The method follows a two-stage process: (1) self-supervised pre-training using contrastive learning with one of three example selection strategies (PolyCL-S, PolyCL-O, or PolyCL-M), where the encoder learns discriminative, task-relevant features from unlabeled data; and (2) supervised fine-tuning, where a decoder is added and the full model is trained on a small set of labeled images to predict segmentation of the target organ(s).}
    \label{fig:polyclarch}
\end{figure*}

Despite the many promising applications of contrastive learning within medical imaging, there are still some challenges. Augmentations often fail to capture the nuanced relationships inherent in medical datasets~\cite{kebaili2023deep} and domain-specific pair generation remains underexplored~\cite{lu2024domain}, limiting the applicability of existing frameworks. Models trained on one imaging modality or anatomical region often fail to generalize to others due to differences in resolution, contrast, and anatomical structures~\cite{tejani2024understanding}. Additionally, the computational demands of modern contrastive frameworks, particularly for large-scale medical datasets, can strain resources in clinical settings, and current methods lack a direct connection to clinical objectives, such as disease-specific localization or quantification, creating a gap between learned representations and practical utility.

\subsection{Medical Image Segmentation}
While contrastive learning largely revolves around pre-training models on large, unlabeled datasets, segmentation models are used to delineate regions of interest, such as tumor margins, from a scan at the pixel level. For years, encoder-decoder architectures inspired by U-Net~\cite{ronneberger2015u} have served as the standard starting point for medical image segmentation models~\cite{azad2024medical}. Such architectures are effective, particularly in their ability to learn spatial hierarchies and recover fine-grained details, but they do suffer from several of the same drawbacks as other deep learning architectures, namely the vanishing gradient problem~\cite{pribadi2024optimization}, where as an architecture deepens, the gradients that are used to update the neural network tend to become smaller and smaller until they vanish as they are backpropagated through the layers of the network. A popular method of addressing this problem is to incorporate residual connections into the U-Net architecture, such as in~\cite{li2024mresunet}. \citet{he2016deep} demonstrated that this approach can improve training efficiency and allow for deeper models by affording better gradient flow across layers. However, while this method does improve performance on complex segmentation tasks, it can still struggle with capturing global context in large medical images, particularly when dealing with long-range dependencies~\cite{pramanik2024daunet}.

More recent medical image segmentation architectures, such as TransUNet~\cite{chen2024transunet} and SwinU-Net~\cite{cao2022swin}, have emerged as strong alternatives to these earlier methods. TransUNet incorporates transformer layers into the U-Net encoder to extract global contexts from images, while the decoder upsamples the encoded features and combines them with high-resolution feature maps to enable precise localization~\cite{chen2021transunet}. This architecture is particularly useful for tasks where the spatial relationships between regions can hold critical diagnostic information, such as organ boundary delineation or lesion segmentation~\cite{chen2024transunet}. Similarly, SwinU-Net builds upon the Swin Transformer~\cite{liu2021swin}, introducing a hierarchical attention mechanism that can efficiently model both global and local features while maintaining computation efficiency. It also uses a window-based attention scheme for better scalability to larger medical images. Despite the power of these transformer-based medical image segmentation models, they do carry with them the downside of high computational complexity and increased resource allocation.

\begin{figure*}
    \centering
    \includegraphics[width=\linewidth, trim={0cm 0.1cm 0cm 0cm},clip]{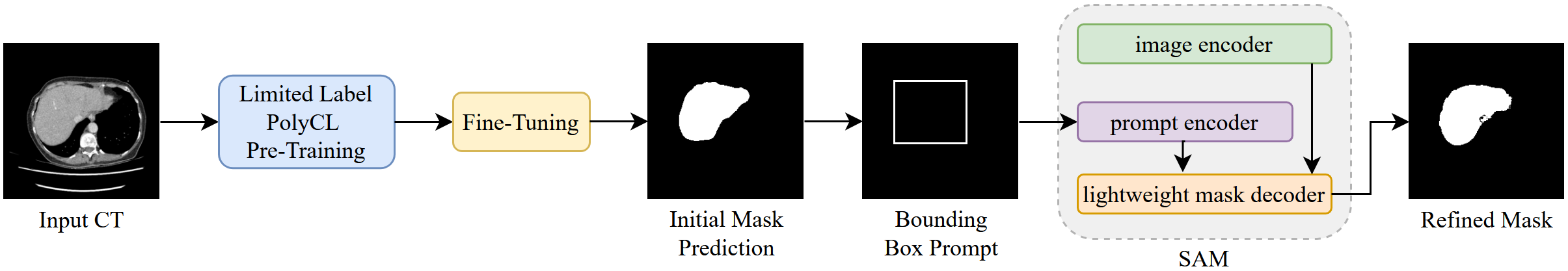}
    \caption{Overview of the SAM-based mask refinement process integrated with PolyCL. Coarse segmentation masks produced by the fine-tuned PolyCL model are first converted into bounding box prompts. These prompts, along with the corresponding CT slices, are then passed into SAM, which refines the masks using a combination of sparse prompt embeddings and dense image embeddings to generate anatomically accurate final segmentation outputs.}
    \label{fig:polyclsam1}
\end{figure*}

Promptable segmentation models such as the Segment Anything Model (SAM) introduced by \citet{kirillov2023segment} offer a promising avenue for avoiding the high computational requirements of other transformer-based methods, due to their strong zero-shot and few-shot generalizability. SAM supports a variety of different prompt formats, including points~\cite{xu2023sppnet}, bounding boxes~\cite{rahman2024pp}, text~\cite{zhang2024evf}, and masks~\cite{xie2024masksam}.  However, manual/semi-automated prompting for SAM carries several of the same flaws as the manual annotation of training data, in that they can be time-consuming to generate~\cite{ward2025annotation}. A potential solution for the issues present in manual or semi-automated prompting of SAM lies in the incorporation of contrastive learning techniques. 

\section{Methods}
\label{sec:methods}
To formulate the problem, we assume an unknown data distribution $p(X, Y)$ over images $X$ and segmentation labels $Y$. 
As shown in Fig.~\ref{fig:polyclarch}, our proposed training framework, PolyCL, uses a two-stage training approach. We first employ self-supervised contrastive learning through our \emph{innovative} sample selection strategy (see Fig.~\ref{fig:updated_teaser}) for pre-training on unlabeled data. Therefore, during pre-training, we assume to have access to $p(X)$ marginalizing out $Y$. We then add a decoder and fine-tune the entire model on small labeled data for the downstream segmentation task. 

\subsection{PolyCL Pre-Training}
The pre-training process begins by using one of our three \textit{novel} example selection strategies. For each anchor slice, $s \in X$, a positive example, $s^+ \in X$, and a negative example, $s^- \in X$, are selected. Because of the limited medical data availability, we devise an organ-based and a scan-based strategy for selecting examples, each requiring a different level of information.

\subsubsection{Organ-based example selection (PolyCL-O):} PolyCL-O requires the knowledge of which slices contain the target organ in the dataset. For reference, Fig.~\ref{fig:liverappearance} depicts CT slices that do and do not contain the target organ (e.g., liver). If the anchor slice contains the target organ, its positive example will also contain the target organ, while its negative example will not. The opposite is true for anchor slices that do not contain the target organ. By choosing examples in this manner, the encoder learns how to represent the target structure before seeing fully annotated data, improving its performance in the actual downstream task. In addition, random selection over all CT scans in the dataset teaches the model inter-scan invariance. Mathematically, this process can be expressed as:

\begin{equation}
\begin{aligned}
s^+ \in \{ x \in X \mid T(x) = T(s) \}\\
s^- \in \{ x \in X \mid T(x) \neq T(s) \},
\end{aligned}
\end{equation}
where $x$ is an arbitrary slice from the dataset $X$ and $T(s)$ is a binary function indicating whether the slice contains the target organ ($T(s) = 1$) or not ($T(s) = 0$).

\subsubsection{Scan-based example selection (PolyCL-S):} PolyCL-S, on the other hand, requires no additional information. For each slice in the dataset, a positive example is selected randomly from the same scan, and a negative example is selected from any scan different from the anchor. This process teaches the encoder intra-scan relationships and enables an understanding of the images even without knowledge of the target structure. Mathematically, it can be expressed as: 

\begin{equation}
\begin{aligned}
s^+ \in \mathcal{S}(s) \setminus \{s\}\\
s^- \in \mathcal{S}^c(s),
\end{aligned}
\end{equation}
where $\mathcal{S}(s)$ is the set of all slices belonging to the same scan as $s$, and $\mathcal{S}^c(s)$ are slices belonging to any other scans.

\begin{figure*}
    \centering
    \includegraphics[width=0.7\linewidth, trim={0cm 0.0cm 0cm 0cm},clip]{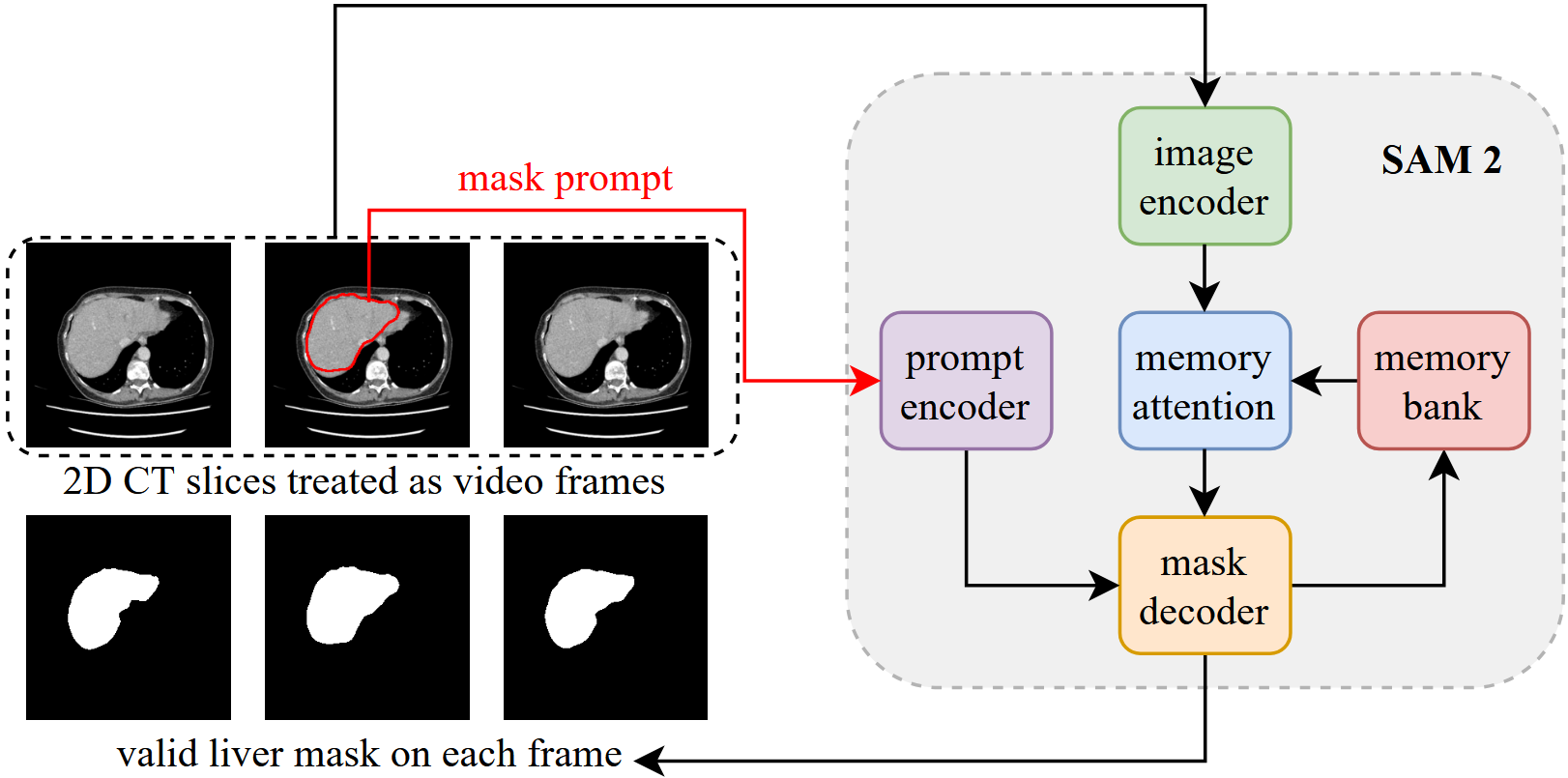}
    \caption{Illustration of the SAM 2-based mask propagation framework integrated with PolyCL. A single annotated reference slice from a 3D CT volume is used to initialize SAM 2. The model then propagates segmentation predictions slice-by-slice through the volume, leveraging spatio-temporal consistency to generate accurate 3D segmentations without requiring additional annotations.}
    \label{fig:polyclsam2}
\end{figure*}

\subsubsection{Mixed-method example selection (PolyCL-M):} The third method, PolyCL-M, combines the organ-based example selection of PolyCL-O with the scan-based approach of PolyCL-S. Similar to PolyCL-O, if an anchor contains the target organ, its positive example will also contain said organ. The opposite is also true, where if an anchor slice does not contain the target organ, so too will its positive example. However, in PolyCL-M, there is an additional criterion in that a positive example must also come from the same scan as the anchor, regardless of the organ information, resembling PolyCL-S. This example selection process teaches the encoder both inter-scan invariance and intra-scan coherence, while also helping the model to better discriminate between similar but contextually different examples. This process can be expressed mathematically as:

\begin{equation}
\begin{aligned}
s^+ \in \{ x \in \mathcal{S}(s) \mid T(x) = T(s) \}\\
s^- \in \{ x \in \mathcal{S}(s) \mid T(x) \neq T(s) \},
\end{aligned}
\end{equation}

\subsubsection{Pretraining Objective}
After example selection is completed for each slice in the dataset, we use the encoder $f$ with projection head $g$ to obtain embedding vectors for the anchor slice, $z = g(f(s))$, and its positive and negative examples, $z^+ = g(f(s^+))$ and $z^- = g(f(s^-))$. The contrastive loss is calculated to enforce similar embeddings between the anchor and its positive example, and dissimilar embeddings between the anchor and its negative example.
\begin{equation}
\label{eq:loss}
L_C = -log\frac{exp(sim(z, z^+) / \tau)}{exp(sim(z, z^+) / \tau) + exp(sim(z, z^-) / \tau)},
\end{equation}
where, $sim(a, b)$ denotes cosine similarity between $a$ and $b$ $(\frac{a \cdot b}{|a| \cdot |b|})$, $M$ is the minibatch size, and the temperature parameter $\tau$ is set to $1/M$. 

\subsection{Fine-Tuning}
After pre-training, the projection head $g$ is discarded from the contrastive learning, and the pretrained encoder is used as the feature extractor. We add a decoder mirroring the encoder with U-Net style skip connections to facilitate pixel-level semantic segmentation. We then fine-tune the entire model on the target segmentation task using a Dice-based loss function.
\begin{equation}
\label{eq:dice}
    L_{Dice}(\hat{Y}, Y) = 1 - \frac{2|\hat{Y} \cap Y|}{|\hat{Y}| + |Y|}.
\end{equation}
where $\hat{Y}$ is the segmentation mask predicted by the model.

\subsection{SAM Integration}
Given that one of the main goals of the PolyCL framework is to achieve strong segmentation performance with limited labeled data, it is natural to investigate ways to refine segmentation predictions further after a rough mask has been generated from one of the PolyCL variants after limited-label fine-tuning. To this end, we explore two methods of improving segmentation results after the pre-training and fine-tuning phases by leveraging the prompting capabilities of SAM. 
\begin{itemize}
    \item First, applying SAM-refinement as a post-processing technique
    \item Second, leveraging the video segmentation capabilities of SAM 2 as a method to segment an entire 3D CT scan based on a single 2D slice input
\end{itemize}

\subsubsection{Mask Refinement}
To explore the capabilities of SAM to revise segmentation predictions after limited-label fine-tuning with PolyCL, we create bounding box prompts for SAM by calculating bounding boxes that define a region of interest (ROI) around each coarse segmentation prediction. Specifically, for each segmented area, the bounding box $bbox$ is calculated as

\begin{equation}
    bbox = (x_{\text{min}}, y_{\text{min}}, x_{\text{max}}, y_{\text{max}}),    
\end{equation}
where
\begin{equation}
\begin{aligned}
    x_{\text{min}} &= \min \{ x_{coord} \mid b(x_{coord}, y_{coord}) = 1 \} \\
    x_{\text{max}} &= \max \{ x_{coord} \mid b(x_{coord}, y_{coord}) = 1 \} \\
    y_{\text{min}} &= \min \{ y_{coord} \mid b(x_{coord}, y_{coord}) = 1 \} \\
    y_{\text{max}} &= \max \{ y_{coord} \mid b(x_{coord}, y_{coord}) = 1 \}.
\end{aligned}
\end{equation}

This bounding box captures the minimum and maximum coordinates along both spatial dimensions, creating a rectangular region that encloses the roughly segmented area in a CT slice.

After the box prompts are generated, they are passed to SAM, where the general segmentation process is as follows: The image encoder from SAM produces feature embeddings capturing important spatial information for the segmentation task. The prompt encoder generates sparse embeddings for ROIs indicated by the created bounding box prompts. The sparse embeddings focus on key points or regions within the image. On the other hand, the dense embeddings via the image encoder provide a broader view, ensuring more comprehensive segmentation coverage in the image. After generating the sparse and dense embeddings, the SAM's decoder is fed these embeddings to predict the final segmentation mask. The predicted masks are refined using the spatial information from the encoder~\cite{ward2025annotation} (see Fig.~\ref{fig:polyclsam1}).

\subsubsection{Mask Propagation}
\label{sec:propagation}
Given a 3D CT volume \( V \in \mathbb{R}^{H \times W \times D} \), where \( H \), \( W \), and \( D \) are the height, width, and number of slices respectively, we aim to produce a full volumetric segmentation by propagating a segmentation mask from a single annotated 2D slice using SAM 2’s video segmentation framework.

First, we define the set of informative slices
\begin{equation}
\mathcal{S} = \{ s \in \{0, \dots, D-1\} \mid \exists (i,j) \, \text{such that} \, Y(i,j,s) = 1 \},
\end{equation}
so that only slices \( s \in \mathcal{S} \) containing non-trivial anatomical structures are considered for propagation. Each slice \( V_s = V[:,:,s] \) is normalized to the range \([0,255]\) and saved as a frame, constructing a pseudo-video \( \mathcal{V} = \{ V_s \mid s \in \mathcal{S} \} \). A reference slice, \( s_r \in \mathcal{S} \), is chosen as the middle slice among those containing annotations.
\begin{equation}
s_r = \mathcal{S}_{\left\lfloor \frac{|\mathcal{S}|}{2} \right\rfloor}.
\end{equation}

Given the binary ground truth mask \( Y_{s_r} \in \{0,1\}^{H \times W} \) for slice \( s_r \), we initialize the SAM 2 predictor by providing \( Y_{s_r} \) as a prompt:
\begin{equation}
\text{State}_0 = \text{Init}(Y_{s_r}, V_{s_r}),
\end{equation}
where \text{State} represents the contents of the memory bank that are updated after each propagation step, \( \text{Init}(\cdot) \) denotes the initialization procedure for SAM 2. Subsequent slice segmentations are obtained by propagating the initial mask through the pseudo-video \(\mathcal{V}\) via the recurrence:
\begin{equation}
\text{State}_{t+1} = \text{Propagate}(\text{State}_t, V_{s_{t+1}}),
\end{equation}
where \( \text{Propagate}(\cdot) \) applies SAM 2’s learned spatio-temporal model to predict the segmentation on the next frame.

For each slice \( s \in \mathcal{S} \), the propagated segmentation mask \( \hat{M}_s \in \{0,1\}^{H \times W} \) is thresholded from the output logits of SAM 2. By propagating from a single 2D mask, this method enables efficient 3D segmentation with minimal manual annotation, providing a promising framework for annotation-efficient clinical deployment (Fig.~\ref{fig:polyclsam2}). 

\section{Experimental Evaluation}
\label{sec:results}

\subsection{Datasets}
We validate our proposed PolyCL method on two separate CT segmentation tasks: liver and kidney segmentations. For liver segmentation, we used three separate datasets at different data settings: Liver Tumor Segmentation (LiTS)~\cite{bilic2023liver}, TotalSegmentator~\cite{Wasserthal_2023}, and Medical Segmentation Decathlon (MSD)~\cite{Antonelli_2022}. To evaluate the performance of PolyCL at segmenting the liver on in-domain data, we used the full 130 abdominal CT scans from the LiTS dataset and split them into training, validation, and testing sets containing 100, 5, and 25 scans, respectively, corresponding to 11,437, 1,139, and 4,827 slices. To test the generalizability of the model, we randomly selected 200 scans (16,300 slices) from the TotalSegmentator dataset and 30 scans (4,567 slices) randomly selected from the MSD dataset.

To evaluate the performance of PolyCL on the kidney segmentation task, we utilized 150 randomly selected CT scans from the 2023 Kidney and Tumor Segmentation (KiTS) Challenge~\cite{heller2023kits21}, split into training, validation, and testing sets consisting of 105, 15, and 30 scans, made up of 23,255, 2,737, and 6,019 slices, respectively. For the generalizability test on this task, we employed 30 CT scans (1,106 slices) from the Beyond the Cranial Vault (BTCV) dataset~\cite{landman2015multi}.

\subsection{Implementation Details}

\textbf{Inputs:} All image slices were 0--1 normalized and reshaped to $256\times 256\times 1$ before passing them to the models. We further preprocessed the images by window-leveling with a width of 400 Hounsfield Units and a center of 40. Since the liver and kidney(s) span a relatively small number of slices in a scan, we extracted only the middle $30\%$ slices to avoid class imbalance.

\noindent\textbf{Model Architecture:} Our PolyCL leverages the encoder-decoder architecture following \cite{haque2021multimix}. We use Leaky ReLU (slope=0.2) and instance normalization between each convolution. We add a global average pool (GAP) at the end of $f$ and a single fully-connected layer as $g$ to generate 256-d embeddings. 

\begin{table}[t]
    \centering
    \caption{Quantitative evaluation demonstrates superior performance of PolyCL in segmenting liver from abdominal CT images in the in-domain LiTS dataset. Mean$\pm$stdev scores are reported by calculating the Dice coefficient and Hausdorff distance after five runs of each of the models.}
    \smallskip
    \resizebox{\linewidth}{!}{%
    \begin{tabular}{l l c c}
    \toprule
         Encoder Backbone & Model & Dice & Hausdorff \\
        \midrule
         - & U-Net &  \textbf{0.897 $\pm$ 0.010} & 13.496 $\pm$ 1.272 \\
         - & nnU-Net & 0.863 $\pm$ 0.034 & 21.407 $\pm$ 1.391 \\
         - & TransUNet & 0.889 $\pm$ 0.005 & 12.68 $\pm$ 2.109 \\
        \midrule
         - & \cite{chaitanya2020contrastive} & 0.840 $\pm$ 0.033 & 8.754 $\pm$ 4.809 \\
         - & \cite{peng2021self} & 0.866 $\pm$ 0.002 & 15.144 $\pm$ 0.099\\
        \midrule
          \multirow{5}{*}{U-Net}  & SupCon & 0.885 $\pm$ 0.005 & 14.709 $\pm$ 0.403 \\
         & SimCLR & 0.892 $\pm$ 0.011 & 13.507 $\pm$ 1.990 \\
        \cmidrule{2-4}
          & PolyCL-S & 0.894 $\pm$ 0.006 & 12.864 $\pm$ 1.942 \\
          & PolyCL-O & 0.888 $\pm$ 0.005 & 11.795 $\pm$ 1.476 \\
          & PolyCL-M & 0.895 $\pm$ 0.003 & 11.690 $\pm$ 0.965 \\
        \midrule
          \multirow{5}{*}{ResU-Net}  & SupCon & 0.888 $\pm$ 0.015 & 13.921 $\pm$ 0.320\\
          & SimCLR & 0.890 $\pm$ 0.013 & 12.936 $\pm$ 1.352 \\
         \cmidrule{2-4}
          & PolyCL-S & 0.894 $\pm$ 0.005 & 12.957 $\pm$ 2.258 \\
          & PolyCL-O & \underline{0.896 $\pm$ 0.006} & \underline{11.664 $\pm$ 0.757} \\
          & PolyCL-M & 0.888 $\pm$ 0.011 & \textbf{11.511 $\pm$ 0.955} \\
        \midrule
         \multirow{5}{*}{TransUNet} & SupCon & 0.885 $\pm$ 0.079 & 15.105 $\pm$ 9.629\\
         & SimCLR & 0.891 $\pm$ 0.021 & 13.394 $\pm$ 1.818 \\
         \cmidrule{2-4}
          & PolyCL-S & 0.896 $\pm$ 0.009 & 12.722 $\pm$ 1.395 \\
         & PolyCL-O & 0.893 $\pm$ 0.006 & 11.835 $\pm$ 1.215\\
         & PolyCL-M & 0.889 $\pm$ 0.004 & 13.370 $\pm$ 11.171 \\
        \bottomrule
    \end{tabular}%
    }
    \label{tab:litsresults}
\end{table}

\noindent\textbf{Baselines:} For baseline comparisons, we used fully-supervised versions of U-Net \cite{ronneberger2015u}, U-Net with residual connections (ResU-Net), nnU-Net \cite{isensee2021nnu}, and TransUNet \cite{chen2024transunet} with random initialization. Additionally, we compared against a ResU-Net pre-trained on ImageNet. We further compared our method to existing contrastive learning frameworks by training and evaluating SupCon \cite{khosla2020supervised} and SimCLR \cite{chen2020simple}, as well as the anatomy-aware contrastive learning methods by \cite{chaitanya2020contrastive} and \cite{peng2021self}. To compare our proposed SAM2 mask refinement method based on PolyCL to existing methods, we evaluate against SAMREFINER \cite{lin2025samrefiner}, which adapts SAM for general-purpose pseudo-label refinement via a multi-prompt excavation strategy and an IoU-adaptive selection mechanism. While SAMRefiner focuses on refining noisy pseudo-masks, our integration of SAM within the PolyCL framework is task-specific and domain-focused for medical image segmentation. Furthermore, our SAM2-based propagation approach leverages spatial continuity across CT slices, which is not possible in SAMREFINER.

\begin{table}[t]
    \centering
    \caption{Quantitative evaluation demonstrates superior performance of PolyCL in segmenting kidney(s) from abdominal CT images in the in-domain KiTS dataset. Mean$\pm$stdev scores are reported by calculating the Dice coefficient and Hausdorff distance after five runs of each of the models.}
    \label{tab:kits}
    \smallskip
    \resizebox{0.9\linewidth}{!}{
    \begin{tabular}{l l c c c}
    \toprule
         Encoder Backbone & Model & Dice & Hausdorff \\
        \midrule
         \multirow{5}{*}{-} & U-Net & \textbf{0.958} & 14.031 \\
          & nnU-Net & 0.916 & 33.126 \\
          & TransUNet & 0.955 & 16.540 \\
        \cmidrule{2-4}
          & \cite{chaitanya2020contrastive} & 0.940 & \textbf{11.248}\\
          & \cite{peng2021self} & 0.932 & 23.487\\
        \midrule
         \multirow{5}{*}{U-Net} & SupCon & 0.953 & 18.036 \\
          & SimCLR & \underline{0.957} & 14.214 \\
         \cmidrule{2-4}
          & PolyCL-S & 0.943 & \underline{12.070} \\
         & PolyCL-O & 0.954 & 15.278 \\
         & PolyCL-M & \underline{0.957} & 20.945 \\
        \bottomrule
    \end{tabular}
    }
\end{table}

\begin{figure}[t]
    \centering
    \resizebox{\linewidth}{!}
    {
    \begin{tabular}{c c c c c}
     U-Net & ResU-Net & PT ResU-Net & nnU-Net & TransU-Net
     \smallskip\\
     \includegraphics[width=0.2\linewidth]{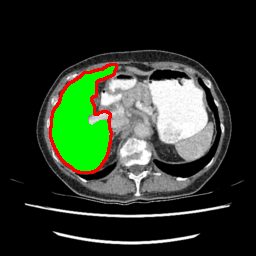}
     &
     \includegraphics[width=0.2\linewidth]{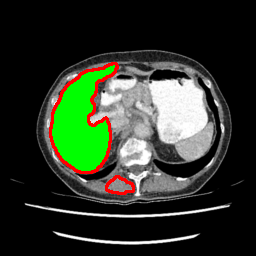}
     &
     \includegraphics[width=0.2\linewidth]{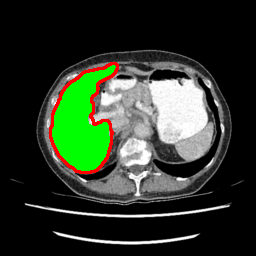}
     &
     \includegraphics[width=0.2\linewidth]{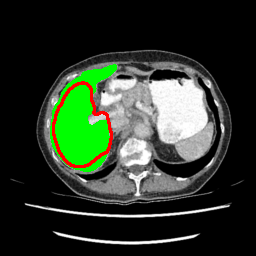}
     &
     \includegraphics[width=0.2\linewidth]{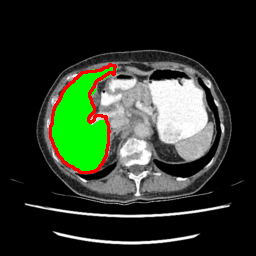}
     \bigskip
     \\
     SupCon & SimCLR & PolyCL-S & PolyCL-O & PolyCl-M
     \smallskip\\
     \includegraphics[width=0.2\linewidth]{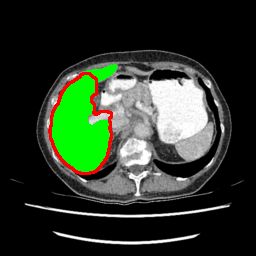}
     &
     \includegraphics[width=0.2\linewidth]{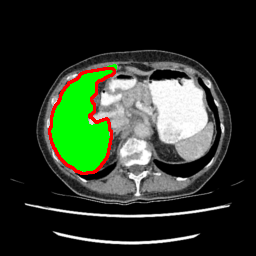}
     &
     \includegraphics[width=0.2\linewidth]{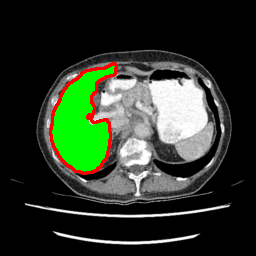}
     &
     \includegraphics[width=0.2\linewidth]{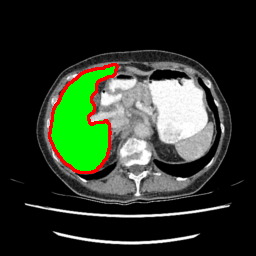}
     &
     \includegraphics[width=0.2\linewidth]{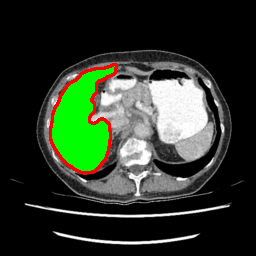}
    \end{tabular}
    }
    \caption{Qualitative comparison demonstrates the superiority of PolyCL over other models in segmenting the liver from the LiTS dataset. Note that PolyCL-based pre-training fixes both over-segmentations (as in the case of U-Net, ResU-Net, and PT ResU-Net) and under-segmentations (present in nnU-Net, TransUNet, SupCon, and SimCLR).}
    \label{fig:masks2}
\end{figure}

\noindent\textbf{Training:} Each model, including each pre-trained encoder, is trained for 100 epochs with a cosine-annealing learning rate scheduler and warm restarts after every 5 epochs. For pre-training the PolyCL variants, we experimented with three different encoder architectures: U-Net, ResU-Net, and TransU-Net. Following pre-training, we fine-tuned the model for 100 epochs. For fine-tuning, the middle 30\% of labeled slices from the CT scans were used.

\noindent\textbf{Hyperparameters:} For pre-training, we used a learning rate of 0.0001 and a batch size of 20, while a learn rate of 0.001 and a batch size of 10 were used for fine-tuning. Each model was trained 5 times to avoid any bias and enhance reliability in model predictions. 

\noindent\textbf{Evaluation:} For evaluation, we use the Dice coefficient as a measure of similarity and Hausdorff distance as the distance metric. The average scores are reported across five iterations of each model.

\subsection{Results}

\subsubsection{Labeled Data Fine-tuning:} 
Our primary findings comparing the proposed model to baseline fully-supervised and contrastive pre-trained methods when segmenting the liver from the LiTS dataset are reported in Table~\ref{tab:litsresults}. We found that the PolyCL variants demonstrate competitive performance against the established supervised baselines, despite not always surpassing the best fully supervised models in absolute terms. Across multiple encoder architectures, PolyCL models consistently match or slightly trail behind the best-performing supervised models in Dice coefficient, while offering notable improvements in Hausdorff distance for several configurations.

Each of the PolyCL variants demonstrates effectiveness, as the best-performing variant is different for each encoder backbone. For the U-Net backbone, PolyCL-M performs the best, surpassing three of the five fully supervised baselines and both self-supervised baselines in terms of Dice coefficient. In this configuration, PolyCL-M also achieves the best Hausdorff distance (2.027\% and 13.454\% lower, respectively, compared to the best-performing fully-supervised and self-supervised baselines), a trend that is reflected in the performance of PolyCL-M with a ResU-Net backbone, which performs the best overall, improving on the best fully-supervised baseline by 3.53\% and the best self-supervised baseline by 11.02\%.

In terms of the Dice coefficient for the ResU-Net backbone, however, PolyCL-O performs the best, outperforming PolyCL-M by 0.8\% and PolyCL-S by 0.2\%. For the TransU-Net encoder backbone, PolyCL-O again performs strongly, achieving the lowest Hausdorff distance by 0.813\% and 11.64\% compared to fully- and self-supervised baselines. In terms of Dice coefficient, PolyCL-S performs the best, outperforming PolyCL-O and PolyCL-M by 0.3\% and 0.7\% respectively. As with the U-Net backbone, all PolyCL variants with a ResU-Net or TransUNet backbone outperform the compared SSL methods with the same backbone.

\begin{figure}[t]
    \centering
    \resizebox{\linewidth}{!}
    {
    \begin{tabular}{c c c c c}
     U-Net & nnU-Net & TransUNet & Chaitanya et al. & Peng et al.
     \smallskip\\
     \includegraphics[width=0.2\linewidth]{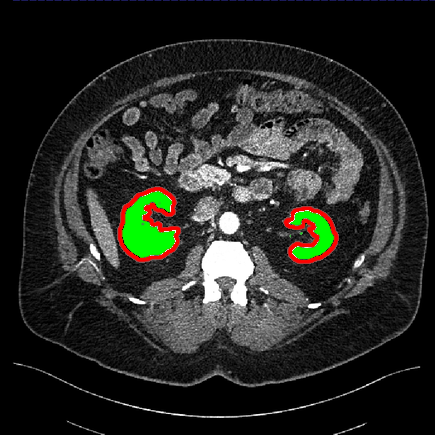}
     &
     \includegraphics[width=0.2\linewidth]{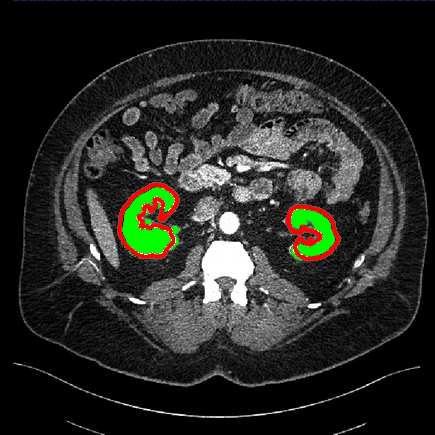}
     &
     \includegraphics[width=0.2\linewidth]{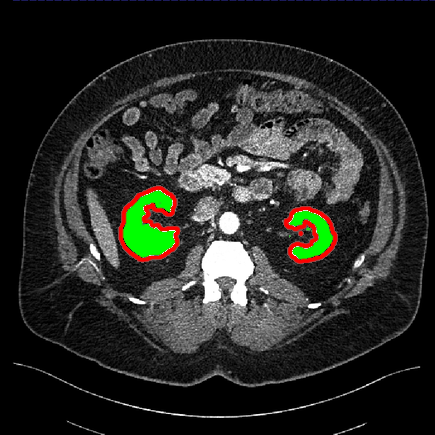}
     &
     \includegraphics[width=0.2\linewidth]{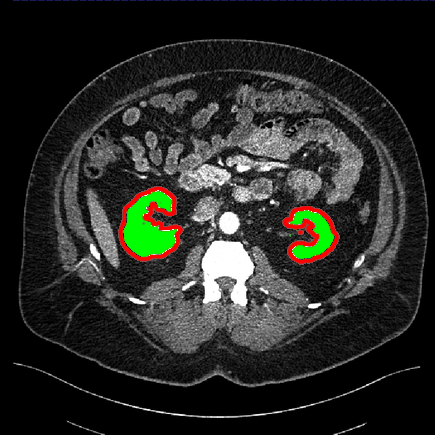}
     &
     \includegraphics[width=0.2\linewidth]{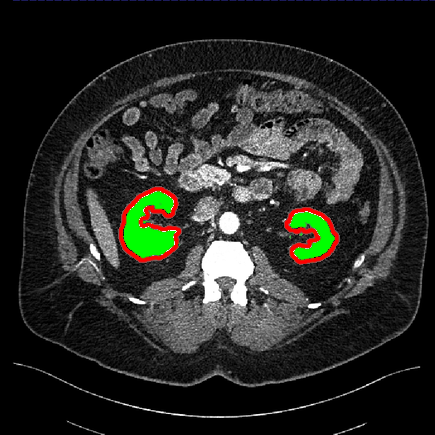}
     \bigskip
     \\
     SupCon & SimCLR & PolyCL-S & PolyCL-O & PolyCl-M
     \smallskip\\
     \includegraphics[width=0.2\linewidth]{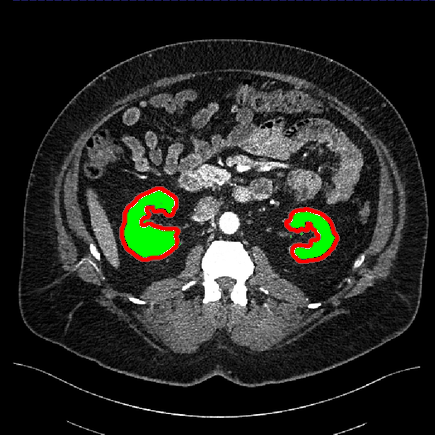}
     &
     \includegraphics[width=0.2\linewidth]{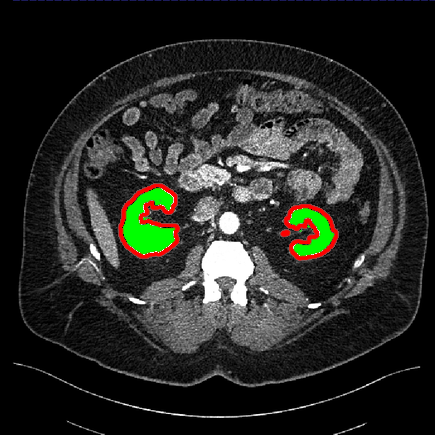}
     &
     \includegraphics[width=0.2\linewidth]{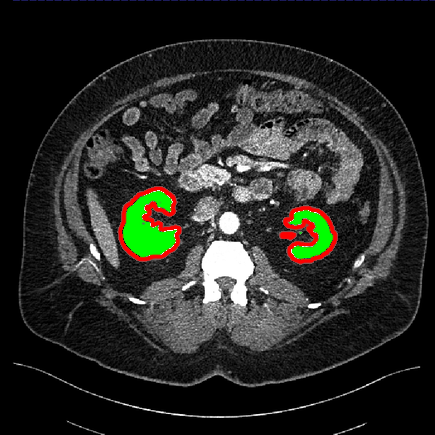}
     &
     \includegraphics[width=0.2\linewidth]{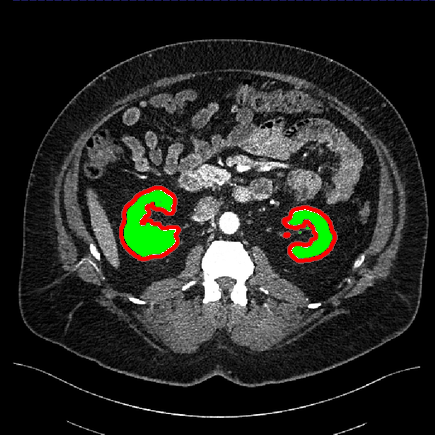}
     &
     \includegraphics[width=0.2\linewidth]{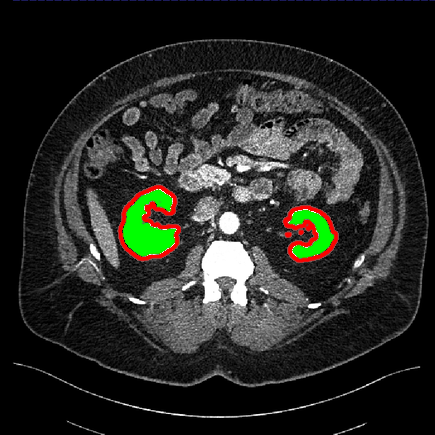}
    \end{tabular}
    }
    \caption{Qualitative comparison demonstrates the superiority of PolyCL over other models in segmenting the kidneys from the KiTS dataset. Note that PolyCL-M is the only model that captures all of the small details present in the ground truth.}
    \label{fig:kitsmasks}
\end{figure}

The results of running PolyCL on the kidney segmentation task (Table~\ref{tab:kits}) remain consistent with the results on liver segmentation. The best-performing PolyCL variant, PolyCL-M, achieves just a 0.001 difference in Dice score compared to the fully-supervised U-Net. In terms of Hausdorff distance, PolyCL-S demonstrates the best performance of the PolyCL variants.

\begin{table}[t]
    \centering
    \caption{Quantitative evaluation demonstrates superior performance of PolyCL in segmenting liver from cross-domain abdominal CT images from the TotalSegmentator dataset. Mean$\pm$stdev scores are reported by calculating the Dice coefficient and Hausdorff distance after five runs of each of the models.}
    \smallskip
    \resizebox{\linewidth}{!}{
    \begin{tabular}{l l c c}
    \toprule
         Encoder Backbone & Model & Dice & Hausdorff \\
        \midrule
         \multirow{7}{*}{-} & U-Net & 0.646 $\pm$ 0.027 & 55.894 $\pm$ 4.063 \\
          & ResU-Net & 0.627 $\pm$ 0.015 & 50.797 $\pm$ 1.855 \\
          & PT ResU-Net & 0.635 $\pm$ 0.025 & 54.856 $\pm$ 6.969 \\
          & nnU-Net & 0.609 $\pm$ 0.025 & 60.301 $\pm$ 3.739 \\
          & TransUNet & 0.604 $\pm$ 0.016 & 73.717 $\pm$ 4.059 \\
        \cmidrule{2-4}
          & \cite{chaitanya2020contrastive} & 0.360 $\pm$ 0.018 & 28.251 $\pm$ 1.044\\
          & \cite{peng2021self} & 0.511 $\pm$ 0.008 & 28.369 $\pm$ 0.263\\
        \midrule
         \multirow{5}{*}{U-Net} & SupCon & 0.612 $\pm$ 0.013 & 57.665 $\pm$ 5.569 \\
          & SimCLR & 0.652 $\pm$ 0.042 & 56.517 $\pm$ 8.703 \\
         \cmidrule{2-4}
         & PolyCL-S & 0.640 $\pm$ 0.029 & 56.179 $\pm$ 6.050 \\
         & PolyCL-O & 0.643 $\pm$ 0.019 & 54.212 $\pm$ 3.854 \\
          & PolyCL-M & 0.633 $\pm$ 0.034 & 55.717 $\pm$ 6.802 \\
        \midrule
         \multirow{5}{*}{ResU-Net} & SupCon & 0.445 $\pm$ 0.061 & 73.557 $\pm$ 5.981 \\
         & SimCLR & 0.590 $\pm$ 0.050 & 61.295 $\pm$ 4.683 \\
         \cmidrule{2-4}
         & PolyCL-S & \textbf{0.753 $\pm$ 0.059} & \textbf{45.463 $\pm$ 8.129} \\
          & PolyCL-O & 0.733 $\pm$ 0.031 & 52.315 $\pm$ 6.559 \\
         & PolyCL-M & 0.720 $\pm$ 0.049 & \underline{49.320 $\pm$ 3.256} \\
        \midrule
         \multirow{5}{*}{TransUNet} & SupCon & 0.467 $\pm$ 0.016 & 88.845 $\pm$ 3.334 \\
         & SimCLR & 0.603 $\pm$ 0.028 & 72.919 $\pm$ 4.653 \\
         \cmidrule{2-4}
         & PolyCL-S & \underline{0.749 $\pm$ 0.016} & 56.065 $\pm$ 6.239 \\
          & PolyCL-O & 0.727 $\pm$ 0.015 & 57.397 $\pm$ 2.197  \\
         & PolyCL-M & 0.740 $\pm$ 0.017 & 57.518 $\pm$ 4.204 \\
        \bottomrule
    \end{tabular}
    }
    \label{tab:tsresults}
\end{table}

\subsubsection{Statistical Analyses:}

To assess whether the observed improvements of PolyCL variants over their corresponding baselines were statistically significant, we performed one-tailed paired t-tests on Dice and Hausdorff scores for all PolyCL variants and encoder backbones against the baseline methods. For models with the U-Net encoder backbone, while many comparisons did not yield statistical significance in Dice coefficient, notable exceptions include the performance of PolyCL against the SupCon baseline (p $<$ 0.01 for all variants). Also, for the Dice coefficient, PolyCL-O demonstrates a statistically significant improvement over TransUNet (p $<$0.05).

For Hausdorff distance, several PolyCL models achieved significant improvements, most prominently PolyCL-O (p $<$0.01) and PolyCL-M (p $<$0.01) over the vanilla U-Net, then again over SupCon (p $<$0.01 for both PolyCL-O and PolyCL-M). PolyCL-O demonstrates a statistically significant improvement over TransUNet for the Hausdorff distance (p $<$0.05), and all three PolyCL variants achieve a statistically significant improvement in Hausdorff distance over nnU-Net (p $<$ 0.01 for all PolyCL variants). These results suggest that, although PolyCL’s gains in Dice score are often modest and sometimes not statistically significant, the method more consistently yields statistically significant reductions in boundary errors, particularly for the organ-based (PolyCL-O) and mixed (PolyCL-M) selection strategies.

\subsubsection{Model Generalization:} To evaluate the generalization capability of PolyCL beyond the in-domain LiTS dataset, we conducted cross-domain experiments using two external datasets: the TotalSegmentator and MSD datasets. These experiments involve fine-tuning models pre-trained on LiTS and directly evaluating them on the new domains without additional domain-specific retraining.

\begin{table}[t]
    \centering
    \caption{Quantitative evaluation demonstrates superior performance of PolyCL in segmenting liver from cross-domain abdominal CT images from the MSD dataset. Mean$\pm$stdev scores are reported by calculating the Dice coefficient and Hausdorff distance after five runs of each of the models.}
    \smallskip
    \resizebox{\linewidth}{!}{
    \begin{tabular}{l l c c}
    \toprule
         Encoder Backbone & Model & Dice & Hausdorff \\
        \midrule
         \multirow{7}{*}{-} & U-Net & 0.746 $\pm$ 0.037 & 46.461 $\pm$ 7.327 \\
          & ResU-Net & 0.737 $\pm$ 0.035 & \textbf{42.485 $\pm$ 2.982} \\
          & PT ResU-Net & 0.747 $\pm$ 0.035 & 45.897 $\pm$ 11.197 \\
          & nnU-Net & 0.730 $\pm$ 0.031 & 51.164 $\pm$ 5.356 \\
          & TransUNet & 0.709 $\pm$ 0.041 & 61.568 $\pm$ 6.188 \\
        \cmidrule{2-4}
         & \cite{chaitanya2020contrastive} & 0.465 $\pm$ 0.016 & 23.042 $\pm$ 2.456\\
          & \cite{peng2021self} & 0.763 $\pm$ 0.007 & 18.450 $\pm$ 0.279 \\
        \midrule
         \multirow{5}{*}{U-Net} & SupCon & 0.659 $\pm$ 0.081 & 51.818 $\pm$ 8.043 \\
         & SimCLR & 0.699 $\pm$ 0.057 & 49.575 $\pm$ 9.635 \\
         \cmidrule{2-4}
         & PolyCL-S & 0.741 $\pm$ 0.035 & 48.080 $\pm$ 9.857 \\
         & PolyCL-O & 0.753 $\pm$ 0.017 & \underline{45.244 $\pm$ 7.731} \\
         & PolyCL-M & 0.724 $\pm$ 0.048 & 48.673 $\pm$ 10.540 \\
        \midrule
        \multirow{5}{*}{ResU-Net} & SupCon & 0.585 $\pm$ 0.005 & 59.209 $\pm$ 4.200 \\
          & SimCLR & 0.671 $\pm$ 0.041 & 54.660 $\pm$ 4.302 \\
         \cmidrule{2-4}
         & PolyCL-S & 0.764 $\pm$ 0.026 & 49.221 $\pm$ 3.310 \\
         & PolyCL-O & 0.741 $\pm$ 0.046 & 53.462 $\pm$ 5.525 \\
         & PolyCL-M & \underline{0.765 $\pm$ 0.015} & 47.650 $\pm$ 3.969 \\
        \midrule
         \multirow{5}{*}{TransUNet} & SupCon & 0.650 $\pm$ 0.013 & 56.881 $\pm$ 12.031 \\
         & SimCLR & 0.705 $\pm$ 0.033 & 54.560 $\pm$ 3.613 \\
         \cmidrule{2-4}
         & PolyCL-S & 0.751 $\pm$ 0.027 & 57.606 $\pm$ 13.632 \\
         & PolyCL-O & 0.761 $\pm$ 0.019 & 53.884 $\pm$ 3.494 \\
         & PolyCL-M & \textbf{0.767 $\pm$ 0.014} & 54.551 $\pm$ 2.131 \\
        \bottomrule\\
    \end{tabular}
    }
    \label{tab:msdresults}
\end{table}

On the TotalSegmentator dataset (Table~\ref{tab:tsresults}), PolyCL variants exhibit clear advantages over both full-supervised baselines and conventional SSL approaches. Notably, the best performing PolyCL variant (PolyCL-S with a ResU-Net encoder backbone) achieves a 10.7\% improvement in Dice coefficient over the best fully-supervised model (U-Net) and a 10.1\% improvement over the best SSL method (SimCLR w/ U-Net backbone). In terms of Hausdorff distance, PolyCL-S improves over the best fully-supervised and self-supervised baselines (ResU-Net and SimCLR w/ U-Net backbone) by a margin of 10.50\% and 19.56\%, respectively.

When evaluated on the MSD dataset (Table~\ref{tab:msdresults}), a similar pattern emerges: PolyCL variants consistently outperform supervised models and alternative SSL strategies. In particular, PolyCL-M with a TransUNet backbone offers a 2\% improvement in Dice coefficient over ImageNet-pretrained ResU-Net, and a 6.8\% improvement over the best SSL method, SimCLR with a U-Net backbone. ]

As reported in Table~\ref{tab:btcvresults}, PolyCL consistently performs well even on the out-of-domain BTCV dataset, when the models were trained on KiTS. PolyCL-S marginally outperforms all other models in terms of Dice score, and PolyCL-M matches the performance of the fully-supervised U-Net and achieves the second-lowest Hausdorff distance.

These results collectively demonstrate that PolyCL pretraining confers a strong advantage in generalization to unseen domains, an increasingly critical capability in medical imaging applications where data distribution shifts are inevitable. Despite being trained solely on LiTS/KiTS data, PolyCL-based models maintain robust performance on datasets that vary significantly in acquisition protocols, anatomical variability, and image characteristics. Moreover, the improvements offered by PolyCL persist across multiple backbone architectures, underscoring the framework's general applicability. By explicitly encouraging intra- and inter-scan relationships during pretraining, PolyCL enables models to learn transferable, task-relevant features that extend well beyond the source domain.

\begin{table}[t]
    \centering
    \caption{Quantitative evaluation demonstrates superior performance of PolyCL in segmenting kidney(s) from abdominal CT images in the cross-domain BTCV dataset. Mean$\pm$stdev scores are reported by calculating the Dice coefficient and Hausdorff distance after five runs of each of the models.}
    \label{tab:btcvresults}
    \smallskip
    \resizebox{0.9\linewidth}{!}{
    \begin{tabular}{l l c c c}
    \toprule
         Encoder Backbone & Model & Dice & Hausdorff \\
        \midrule
         \multirow{5}{*}{-} & U-Net & \underline{0.491} & 53.250 \\
          & nnU-Net & 0.465 & 57.395 \\
          & TransUNet & 0.452 & 68.305 \\
        \cmidrule{2-4}
          & \cite{chaitanya2020contrastive} & 0.308 & \textbf{26.309}\\
          & \cite{peng2021self} & 0.432 & 54.009\\
        \midrule
         \multirow{5}{*}{U-Net} & SupCon & 0.466 & 56.404 \\
          & SimCLR & 0.471 & 54.773 \\
         \cmidrule{2-4}
         & PolyCL-S & \textbf{0.492} & 53.165 \\
         & PolyCL-O & 0.470 & 55.029 \\
         & PolyCL-M & \underline{0.491} & \underline{53.161} \\
        \bottomrule\\
    \end{tabular}
    }
\end{table}

\begin{table*}[t]
    \centering
    \caption{Dice Scores comparing PolyCL variants with and without SAM. Mean$\pm$stdev are reported.}
    \smallskip
    \small
    \resizebox{0.8\linewidth}{!}{
    \begin{tabular}{l c c c c c}
    \toprule
         Model & 5\% & 10\% & 20\% & 30\% & 50\% \\
    \midrule
         PolyCL-S & 0.489 $\pm$ 0.021 & 0.469 $\pm$ 0.025 & 0.480 $\pm$ 0.022 & 0.436 $\pm$ 0.017 & 0.479 $\pm$ 0.019 \\
         PolyCL-O & 0.482 $\pm$ 0.027 & 0.504 $\pm$ 0.045 & 0.568 $\pm$ 0.021 & 0.579 $\pm$ 0.009 & 0.583 $\pm$ 0.007 \\
         PolyCL-M & 0.495 $\pm$ 0.015 & 0.520 $\pm$ 0.012 & 0.564 $\pm$ 0.015 & 0.582 $\pm$ 0.013 & 0.638 $\pm$ 0.070 \\
         SAMRefiner-S & 0.613 $\pm$ 0.061 & 0.702 $\pm$ 0.028 & 0.590 $\pm$ 0.041 & 0.547 $\pm$ 0.027 & 0.612 $\pm$ 0.026 \\
         SAMRefiner-O &  0.658 $\pm$ 0.035 & 0.661 $\pm$ 0.089 & 0.721 $\pm$ 0.029 & 0.785 $\pm$ 0.027 & 0.806 $\pm$ 0.019 \\
         SAMRefiner-M & 0.642 $\pm$ 0.030 & 0.702 $\pm$ 0.028 & 0.732 $\pm$ 0.024 & 0.785 $\pm$ 0.020 & 0.819 $\pm$ 0.016 \\
         PolyCL-S+SAM & 0.616 $\pm$ 0.130 & 0.647 $\pm$ 0.034 & 0.662 $\pm$ 0.031 & 0.614 $\pm$ 0.029 & 0.660 $\pm$ 0.026 \\
         PolyCL-O+SAM & \underline{0.664 $\pm$ 0.037} & \underline{0.693 $\pm$ 0.060} & \underline{0.772 $\pm$ 0.015} & \underline{0.793 $\pm$ 0.009} & \underline{0.805 $\pm$ 0.009} \\
         PolyCL-M+SAM & \textbf{0.682 $\pm$ 0.021} & \textbf{0.718 $\pm$ 0.017} & \textbf{0.780 $\pm$ 0.021} & \textbf{0.802 $\pm$ 0.017} & \textbf{0.825 $\pm$ 0.009} \\
    \bottomrule
    \end{tabular}
    }
    \label{tab:dice_sam}\
\end{table*}

\begin{table*}[t]
    \centering
    \caption{Hausdorff Distances comparing PolyCL variants with and without SAM. Mean$\pm$stdev are reported.}
    \smallskip
    \small
    \resizebox{0.8\linewidth}{!}{
    \begin{tabular}{l c c c c c}
    \toprule
         Model & 5\% & 10\% & 20\% & 30\% & 50\% \\
    \midrule
         PolyCL-S & 65.472 $\pm$ 9.061 & 76.991 $\pm$ 9.007 & 68.293 $\pm$ 13.927 & 82.151 $\pm$ 7.679 & 71.557 $\pm$ 7.035 \\
         PolyCL-O & 66.081 $\pm$ 4.417 & 65.614 $\pm$ 11.605 & 45.960 $\pm$ 3.579 & 42.220 $\pm$ 9.141 & 36.046 $\pm$ 3.892 \\
         PolyCL-M & 70.471 $\pm$ 5.735 & 63.148 $\pm$ 4.668 & 51.280 $\pm$ 4.982 & 41.706 $\pm$ 5.562 & 22.094 $\pm$ 3.318 \\
         SAMRefiner-S & 18.228 $\pm$ 1.403 & 19.463 $\pm$ 2.605 & 21.037 $\pm$ 1.247 & 22.555 $\pm$ 0.896 & 19.382 $\pm$ 1.516 \\
         SAMRefiner-O &  17.691 $\pm$ 1.540 & 16.859 $\pm$ 3.855 & 14.216 $\pm$ 1.873 & 11.207 $\pm$ 2.155 & 9.473 $\pm$ 1.428 \\
         SAMRefiner-M & 19.297 $\pm$ 2.468 & 19.463 $\pm$ 2.605 & 15.128 $\pm$ 1.936 & 11.466 $\pm$ 0.020 & 9.084 $\pm$ 1.237 \\
         PolyCL-S+SAM & 5.813 $\pm$ 0.677 & 5.762 $\pm$ 0.200 & 5.665 $\pm$ 0.292 & 6.104 $\pm$ 0.325 & \underline{5.681 $\pm$ 0.200} \\
         PolyCL-O+SAM & \underline{5.608 $\pm$ 0.451} & \underline{5.397 $\pm$ 0.615} & \underline{4.763 $\pm$ 0.166} & \textbf{4.656 $\pm$ 0.100} & \textbf{4.695 $\pm$ 0.093} \\
         PolyCL-M+SAM & \textbf{5.484 $\pm$ 0.191} & \textbf{5.148 $\pm$ 0.156} & \textbf{4.761 $\pm$ 0.148} & \underline{4.668 $\pm$ 0.166} & 8.107 $\pm$ 7.396 \\
    \bottomrule
    \end{tabular}
    }
    \label{tab:hd_sam}
\end{table*}

\begin{table*}[t]
    \centering
    \caption{Dice Scores comparing PolyCL variants with and without SAM 2. Mean$\pm$stdev are reported.}
    \smallskip
    \small
    \resizebox{0.85\linewidth}{!}{
    \begin{tabular}{l c c c c c}
    \toprule
         Model & 5\% & 10\% & 20\% & 30\% & 50\% \\
    \midrule
         PolyCL-S & 0.489 $\pm$ 0.021 & 0.469 $\pm$ 0.025 & 0.480 $\pm$ 0.022 & 0.436 $\pm$ 0.017 & 0.479 $\pm$ 0.019 \\
         PolyCL-O & 0.482 $\pm$ 0.027 & 0.504 $\pm$ 0.045 & 0.568 $\pm$ 0.021 & 0.579 $\pm$ 0.009 & 0.583 $\pm$ 0.007 \\
         PolyCL-M & 0.495 $\pm$ 0.015 & 0.520 $\pm$ 0.012 & 0.564 $\pm$ 0.015 & 0.582 $\pm$ 0.013 & 0.638 $\pm$ 0.070 \\
         PolyCL-S+SAM 2 & 0.597 $\pm$ 0.185 & 0.627 $\pm$ 0.048 & 0.642 $\pm$ 0.044 & 0.595 $\pm$ 0.041 & 0.640 $\pm$ 0.037 \\
         PolyCL-O+SAM 2 & 0.622 $\pm$ 0.160 & 0.649 $\pm$ 0.260 & 0.723 $\pm$ 0.065 & 0.743 $\pm$ 0.039 & 0.754 $\pm$ 0.039 \\
         PolyCL-M+SAM 2 & 0.632 $\pm$ 0.037 & 0.666 $\pm$ 0.030 & 0.723 $\pm$ 0.037 & 0.744 $\pm$ 0.030 & 0.765 $\pm$ 0.016 \\
    \bottomrule
    \end{tabular}
    }
    \label{tab:dice_sam2}
\end{table*}

\begin{table*}[t]
    \centering
    \caption{Hausdorff Distances comparing PolyCL variants with and without SAM 2. Mean$\pm$stdev are reported.}
    \smallskip
    \small
    \resizebox{0.85\linewidth}{!}{
    \begin{tabular}{l c c c c c}
    \toprule
         Model & 5\% & 10\% & 20\% & 30\% & 50\% \\
    \midrule
         PolyCL-S & 65.472 $\pm$ 9.061 & 76.991 $\pm$ 9.007 & 68.293 $\pm$ 13.927 & 82.151 $\pm$ 7.679 & 71.557 $\pm$ 7.035 \\
         PolyCL-O & 66.081 $\pm$ 4.417 & 65.614 $\pm$ 11.605 & 45.960 $\pm$ 3.579 & 42.220 $\pm$ 9.141 & 36.046 $\pm$ 3.892 \\
         PolyCL-M        & 70.471 $\pm$ 5.735 & 63.148 $\pm$ 4.668 & 51.280 $\pm$ 4.982 & 41.706 $\pm$ 5.562 & 22.094 $\pm$ 3.318 \\
         PolyCL-S+SAM 2  & 75.993 $\pm$ 46.116 & 75.361 $\pm$ 13.575 & 74.149 $\pm$ 19.592 & 80.070 $\pm$ 22.136 & 74.421 $\pm$ 13.575 \\
         PolyCL-O+SAM 2 & 38.632 $\pm$ 6.626 & 37.209 $\pm$ 9.032 & 32.797 $\pm$ 2.448 & 31.990 $\pm$ 1.472 & 32.333 $\pm$ 1.369 \\
         PolyCL-M+SAM 2 & 22.258 $\pm$ 0.119 & 20.910 $\pm$ 0.097 & 19.345 $\pm$ 0.092 & 18.937 $\pm$ 0.104 & 32.899 $\pm$ 4.611 \\
    \bottomrule
    \end{tabular}
    }
    \label{tab:hd_sam2}
\end{table*}

\subsubsection{Reduced-label Fine-tuning and Mask Refinement}
To further assess the data efficiency of PolyCL, we conducted a reduced-label fine-tuning study on the LiTS dataset, varying the proportion of labeled data available during training. The performance of all PolyCL variants trained with a U-Net encoder across all label fractions (5\% to 50\%) is shown in Table~\ref{tab:dice_sam} and Table~\ref{tab:hd_sam}. Here, the superiority of PolyCL-M over the -S and -O variants is apparent, as it achieves the highest Dice coefficient across four of the five reduced label settings, and the lowest Hausdorff distance across three of the five.

Across all label fractions, all three PolyCL variants exhibited large performance gains when refined with SAM. For example, PolyCL-M+SAM achieved a Dice score of 0.825 at 50\% labeled data, which is substantially higher than its unrefined counterpart, as well as SAMREFINER, which only achieved a Dice score of 0.819. Even at just 5\% labeled data, PolyCL-M+SAM attained a Dice of 0.682, outperforming its baseline version by nearly 19\% and SAMREFINER by 4\%.

In addition to improvements to the Dice coefficient, SAM refinement dramatically improved boundary accuracy. As shown in Table~\ref{tab:hd_sam}, Hausdorff distances for all PolyCL+SAM combinations dropped sharply compared to the original PolyCL outputs as well as the SAMREFINER-refined samples. For example, the best-performing configuration, PolyCL-O + SAM, achieved a Hausdorff distance of 4.695, indicating precise boundary localization with minimal annotation.

These results highlight two key advantages of the PolyCL framework: (1) strong segmentation performance under severely limited supervision and (2) seamless integration with downstream refinement modules like SAM to further enhance anatomical consistency. By enabling robust segmentation from small labeled subsets and yielding masks with high structural fidelity, PolyCL+SAM presents a compelling approach for potential annotation-efficient deployment in clinical settings.

\begin{figure}
    \centering
    \resizebox{\linewidth}{!}
    {
    \begin{tabular}{c c c c}
      & Dice: 0.924 &  & Dice: 0.965
     \smallskip\\
     \includegraphics[width=0.24\linewidth]{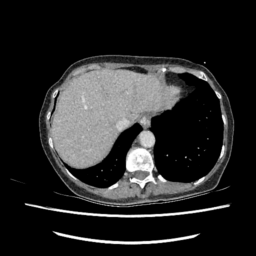}
     &
     \includegraphics[width=0.24\linewidth]{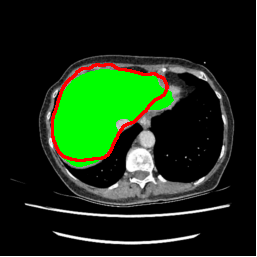}
     &
     \includegraphics[width=0.24\linewidth]{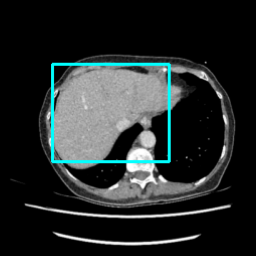}
     &
     \includegraphics[width=0.24\linewidth]{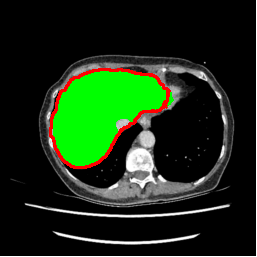}
     \\
     CT Slice & Predicted Mask & Bbox Prompt & SAM-refined Mask
     \\
     \addlinespace[0.5em]
     & Dice: 0.936 &  & Dice: 0.964
     \smallskip\\
     \includegraphics[width=0.24\linewidth]{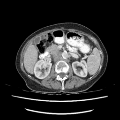}
     &
     \includegraphics[width=0.24\linewidth]{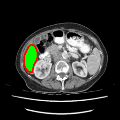}
     &
     \includegraphics[width=0.24\linewidth]{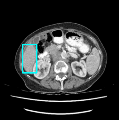}
     &
     \includegraphics[width=0.24\linewidth]{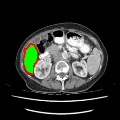}
     \\
     CT Slice & Predicted Mask & Bbox Prompt & SAM-refined Mask
    \end{tabular}
    }
    \caption{Qualitative comparison demonstrating the performance of PolyCL-SAM at refining rough segmentation outputs.}
    \label{fig:masks2}
\end{figure}

\subsubsection{Mask Propagation with SAM 2}

Tables~\ref{tab:dice_sam2} and~\ref{tab:hd_sam2} present the performance of integrating the mask propagation framework for each of the PolyCL variants. We evaluated the models trained with a U-Net encoder under varying labeled training data settings, ranging between 5\%-50\%. Compared to baseline PolyCL variants without any refinement or propagation, SAM 2 yields considerable gains in segmentation performance, especially under severely limited labeled data settings. 
PolyCL-M+SAM 2 achieved consistently high Dice scores across all label fractions, peaking at \( 0.765 \pm 0.016 \) for the 50\% label setting. While slightly lower than PolyCL-M+SAM (Dice \( = 0.825 \pm 0.009 \)), SAM 2 propagation remained highly effective even when initialized from a single annotated slice. Notably, the performance difference between SAM and SAM 2 narrows at lower annotation levels (e.g., Dice \( = 0.682 \) vs. \( 0.632 \) at 5\%), underscoring SAM 2’s utility in ultra-low-label settings.

In terms of Hausdorff distance, SAM 2 also demonstrates strong results. For example, PolyCL-M+SAM 2 achieved a Hausdorff distance of \( 18.937 \pm 0.104 \) at 30\% labeled data, a drastic reduction from the original PolyCL-M model’s \( 41.706 \pm 5.562 \). While SAM refinement still yielded the lowest overall Hausdorff values (e.g., PolyCL-O + SAM reached \( 4.695 \pm 0.093 \)), the propagation-based method significantly outperformed all unrefined models under low supervision.

These results highlight the viability of using a single reference annotation to produce high-quality volumetric segmentations across a 3D scan. When paired with PolyCL’s pretrained representations, the SAM 2 framework offers a compelling strategy for clinical deployment, enabling high-fidelity mask generation with minimal annotation and without requiring slice-wise prompting or retraining.

\subsubsection{Ablation Experiments}
There are many pathways for ablation experiments in this work. On the PolyCL side, we report an analysis of different configurations of batch sizes and temperatures for both pre-training and fine-tuning, as well as different dimensions for the projection head in the contrastive encoder. These results are reported in Table~\ref{tab:polycl_ablation}. 

Because of the large number of possible configurations of these parameters (81), we employ a Taguchi L9~\cite{kv2019overview} design to obtain informative results in a smaller number of experiments (9). From these results, it can be observed that none of the other possible configurations for the parameters beat the setup that we employed for the final version of our model (pre-training batch size of 20, fine-tuning batch size of 10, 256$\times$256 projection head dimensions).

On the SAM side, SAM2 is capable of taking in multiple inputs to be propagated throughout an image stream, thus Table~\ref{tab:sam2_ablation} reports the results of an ablation experiment on the configuration of labeled slices used for this mask propagation. Investigated were the number of 2D slices (1-3) used to propagate through the 3D scan, as well as the location within the volume that these slices were sampled from (the beginning/middle/end of the scan). From this experiment, it is observed that sampling from the middle of the volume achieves the best results across the analyzed percentages of limited labeled data. Additionally, propagating with just a single slice seems to be the most beneficial compared to increasing the number of slices. This is likely due to the fact that the labeled slices come from coarse mask predictions, and if multiple coarse masks that are distinct from each other are sampled, it can be hard for the model to understand realistic patterns between slices.

\begin{table}[t]
    \centering
    \caption{Impact of varying numbers and positions for the labeled slices used for SAM2 mask propagation. Mean$\pm$stdev are reported.}
    \smallskip
    \Huge
    \resizebox{\linewidth}{!}{
    \begin{tabular}{c c l c}
    \toprule
         Labeled Data (\%) & \#Slices4Propagation & Slice-Position & Dice \\
    \midrule
         \multirow{9}{*}{5\%} & \multirow{3}{*}{1} & Beginning & 0.615 $\pm$ 0.336 \\
          &  & Middle & 0.632 $\pm$ 0.037 \\
          &  & End & 0.586 $\pm$ 0.368 \\
          \cmidrule(lr){2-4}
          & \multirow{3}{*}{2} & Beginning & 0.587 $\pm$ 0.385 \\
          &  & Middle & 0.613 $\pm$ 0.391 \\
          &  & End & 0.550 $\pm$ 0.407 \\
          \cmidrule(lr){2-4}
          & \multirow{3}{*}{3} & Beginning & 0.558 $\pm$ 0.407 \\
          &  & Middle & 0.584 $\pm$ 0.356 \\
          &  & End & 0.526 $\pm$ 0.001 \\
    \midrule
         \multirow{9}{*}{50\%} & \multirow{3}{*}{1} & Beginning & 0.751 $\pm$ 0.935 \\
          &  & Middle & 0.765 $\pm$ 0.016 \\
          &  & End & 0.716 $\pm$ 0.385 \\
          \cmidrule(lr){2-4}
          & \multirow{3}{*}{2} & Beginning & 0.767 $\pm$ 0.357 \\
          &  & Middle & 0.782 $\pm$ 0.353 \\
          &  & End & 0.751 $\pm$ 0.354 \\
          \cmidrule(lr){2-4}
          & \multirow{3}{*}{3} & Beginning & 0.627 $\pm$ 0.192 \\
          &  & Middle & 0.683 $\pm$ 0.059 \\
          &  & End & 0.600 $\pm$ 0.362 \\
    \bottomrule
    \end{tabular}
    }
    \label{tab:sam2_ablation}
\end{table}

\section{Discussion}
\label{sec:discussion}

\subsection{Limitations and Future Work}
While PolyCL demonstrates strong performance in both in-domain and cross-domain segmentation tasks, several limitations remain. First, the framework currently focuses primarily on organ-level segmentation using CT data. Although the proposed organ-based, scan-based, and mixed example selection strategies are designed to be generalizable, their performance on smaller, less distinct anatomical targets or in modalities such as MRI or ultrasound remains untested. Second, the integration with SAM relies on accurate coarse mask predictions from PolyCL to generate bounding box prompts. In scenarios where coarse outputs are highly inaccurate, SAM refinement may not yield meaningful improvements. Similarly, SAM 2’s mask propagation assumes high inter-slice anatomical consistency, and performance may degrade in cases of irregular anatomy or the presence of artifacts. Additionally, both SAM refinement and SAM 2 propagation introduce computational overhead, which may limit applicability in resource-constrained clinical environments. Finally, although PolyCL was tested on multiple public datasets, all experiments were retrospective and drawn from open challenges, which may not fully capture the variability and complexity of real-world clinical workflows. Future work will focus on building solutions to these highlighted challenges.

\subsection{Clinical Relevance}
From a deployment perspective, the organ-based selection strategy (PolyCL-O) introduces a modest preprocessing overhead (about 1.652 seconds per CT volume to identify slices containing the target organ). While negligible for small datasets, this could accumulate to several hours for larger datasets involving thousands of scans. However, as demonstrated by our results (Tables 1–9), incorporating organ-level information typically improves segmentation accuracy over scan-level selection alone (PolyCL-S). For use cases where accurate delineation of patient anatomy is critical, such as pre-surgical planning, the additional preprocessing time is likely justified. Furthermore, slice selection is a one-time cost incurred only during training. Once trained, PolyCL-O or PolyCL-M can be applied to unseen data without re-performing organ-based selection.

In terms of the practicality of deploying the SAM-based mask refinement system, we estimate that it takes approximately 50.44 milliseconds to refine one mask prediction. As with the organ-based selection strategy, this represents a marginal computational cost for smaller datasets that could balloon as the size of the dataset increases. However, the time it would take for a radiologist to manually annotate through the slices in a 3D volume is much greater than this, meaning that if SAM refinement were able to achieve even marginally accurate segmentations, this would represent an exponential reduction in human effort spent labeling samples.

\begin{table}[t]
    \centering
    \caption{Impact of various batch sizes, projection head dimensions, and temperatures on the PolyCL architecture.}
    \smallskip
    \small
    \resizebox{\linewidth}{!}{
    \begin{tabular}{c c c c c c}
    \toprule
         Batch Size (PT) & Temperature (PT) & Batch Size (FT) & Temperature (FT) & Projection Head & Dice \\
    \midrule
         \multirow{3}{*}{10} & \multirow{3}{*}{0.100}
            & 10 & 0.100 & 128x128 & 0.869 \\
            &   & 20 & 0.050 & 256x256 & 0.855 \\
            &   & 30 & 0.033 & 512x512 & 0.882 \\
         \midrule
         \multirow{3}{*}{20} & \multirow{3}{*}{0.050}
            & 10 & 0.100 & 128x128 & 0.854 \\
            &   & 20 & 0.050 & 256x256 & 0.897 \\
            &   & 30 & 0.033 & 512x512 & 0.881 \\
         \midrule
         \multirow{3}{*}{30} & \multirow{3}{*}{0.033}
            & 10 & 0.100 & 128x128 & 0.882 \\
            &   & 20 & 0.050 & 256x256 & 0.870 \\
            &   & 30 & 0.033 & 512x512 & 0.886 \\
    \bottomrule
    \end{tabular}
    }
    \label{tab:polycl_ablation}
\end{table}

\section{Conclusions}
\label{sec:conclusion}

We have presented a novel contrastive learning-based self-supervised learning method (PolyCL) employing innovative organ-based and scan-based example selection strategies for medical image segmentation. Evaluating on multiple abdominal CT datasets, we have found that our pre-training strategy generates significant benefits and can use existing data more efficiently than fully supervised and other contrastive learning-based methods. Additionally, PolyCL demonstrated improved generalization when fine-tuned and tested on out-of-distribution data. 

To further improve segmentation performance, particularly under limited annotation, we explored the integration of SAM. We introduced two complementary strategies: SAM refinement, which enhances the structural accuracy of predicted masks using box prompts, and SAM 2 mask propagation, which enables 3D volumetric segmentation from a single annotated slice using SAM 2’s video-based modeling. Both techniques significantly boosted segmentation quality, with SAM refinement yielding precise boundaries and SAM 2 offering robust performance in single-label scenarios. Our future work will focus on extending the framework to additional anatomical structures and imaging modalities.

\needspace{4\baselineskip}
\ethics{The work follows appropriate ethical standards in conducting research and writing the manuscript, following all applicable laws and regulations regarding treatment of animals or human subjects.}

\coi{We declare we don't have conflicts of interest.}

\data{For this work, we trained/evaluated our models on four publicly available datasets: LiTS \cite{bilic2023liver} which is available for download at ~\url{https://competitions.codalab.org/competitions/17094}; TotalSegmentator \cite{Wasserthal_2023}, available for download at ~\url{https://zenodo.org/records/10047292}; MSD \cite{Antonelli_2022}, available for download at ~\url{http://medicaldecathlon.com/dataaws/}; and BTCV, available for download at ~\url{https://www.synapse.org/Synapse:syn3193805/wiki/89480}. The code, along with pre-trained checkpoints and data samples, is available in our public GitHub repository, linked at the end of our abstract.
}

\bibliography{references}

\end{document}